\documentclass[sigconf]{acmart}
\AtBeginDocument{%
  \providecommand\BibTeX{{%
    \normalfont B\kern-0.5em{\scshape i\kern-0.25em b}\kern-0.8em\TeX}}}

\copyrightyear{2022}
\acmYear{2022}
\setcopyright{rightsretained}
\acmConference[ISLPED '22]{ACM/IEEE International Symposium on Low Power Electronics and Design}{August 1--3, 2022}{Boston, MA, USA}
\acmBooktitle{ACM/IEEE International Symposium on Low Power Electronics and Design (ISLPED '22), August 1--3, 2022, Boston, MA, USA}
\acmDOI{10.1145/3531437.3539717}
\acmISBN{978-1-4503-9354-6/22/08}

\usepackage{mathtools}
\DeclarePairedDelimiter{\abs}{\lvert}{\rvert}
\DeclarePairedDelimiter{\norma}{\lVert}{\rVert}

\usepackage{subfig}
\usepackage{enumitem}

\usepackage{tikz}
\newcommand*\circled[1]{\tikz[baseline=(char.base)]{
            \node[shape=circle,draw,inner sep=0.8pt, minimum size=2pt] (char) {#1};}}
\newcommand{\rpoint}[1]{\circled{{\fontfamily{pcr}\selectfont\footnotesize{#1}}}}

\usepackage{fancyhdr,lipsum}
\fancypagestyle{firstpage}{
  \fancyhf{}
  \fancyhead[C]{To appear at the ACM/IEEE International Symposium on Low Power Electronics and Design (ISLPED), August 2022, Boston, MA, USA.}
  \fancyfoot[C]{\thepage}
}

\setcopyright{none}
\renewcommand\footnotetextcopyrightpermission[1]{} 
\begin{document}

\title{Enabling Capsule Networks at the Edge through Approximate Softmax and Squash Operations}

\author{Alberto Marchisio\textsuperscript{1,*}, Beatrice Bussolino\textsuperscript{2,*}, Edoardo Salvati\textsuperscript{2,*}, Maurizio Martina\textsuperscript{2}, Guido Masera\textsuperscript{2}, Muhammad Shafique\textsuperscript{3}}
\affiliation{%
  \textsuperscript{1}
  \institution{\textit{Institute of Computer Engineering, Technische Universit{\"a}t Wien (TU Wien), Vienna, Austria}}
  \country{\textsuperscript{2}\textit{Department of Electronics and Telecommunications, Politecnico di Torino, Turin, Italy}\\
  \textsuperscript{3}\textit{eBrain Lab, Division of Engineering, New York University Abu Dhabi, UAE}}
}
\email{alberto.marchisio@tuwien.ac.at, beatrice.bussolino@polito.it, edoardo.salvati@studenti.polito.it,}
\email{maurizio.martina@polito.it, guido.masera@polito.it, muhammad.shafique@nyu.edu}

\thanks{*These authors contributed equally to this work.}

\renewcommand{\shortauthors}{A. Marchisio, et al.}

\begin{abstract}
Complex Deep Neural Networks such as Capsule Networks (CapsNets) exhibit high learning capabilities at the cost of compute-intensive operations. To enable their deployment on edge devices, we propose to leverage approximate computing for designing approximate variants of the complex operations like softmax and squash. In our experiments, we evaluate tradeoffs between area, power consumption, and critical path delay of the designs implemented with the ASIC design flow, and the accuracy of the quantized CapsNets, compared to the exact functions.
\end{abstract}





\maketitle
\thispagestyle{firstpage}

\section{Introduction}

In recent years, Deep Neural Networks (DNNs) have achieved outstanding performance in a wide range of applications~\cite{Mihai2019DL4AD}\cite{Li2021CNNApplications}. Among the latest DNN models, Capsule Networks (CapsNets)~\cite{shallowcaps} enable high learning capabilities due to the capsules, which add a layer of abstraction compared to the traditional neurons of DNNs. Despite their groundbreaking success, the most advanced DNNs, such as CapsNets, exhibit high complexity due to their compute-intensive operations, which hinders their deployments on energy-constrained edge devices. Therefore, several optimizations have been proposed to increase the performance and reduce the energy consumption of complex DNNs on edge devices, such as network pruning~\cite{Frankle2018LotteryTicketHypothesis}\cite{Han2015Pruning}\cite{Marchisio2018PruNet}, and quantization~\cite{qcaps}\cite{Yiren2018AdaptiveQuantization}. In this work, we focus on leveraging approximate computing for CapsNet optimization. Therefore, our approach is orthogonal to other optimization techniques (e.g., quantization), as we directly perform experiments on the quantized CapsNets.

\subsection{Motivations and Research Challenges}

\begin{figure}[h]
    \centering
    \vspace{-2pt}
    \includegraphics[width=.95\linewidth]{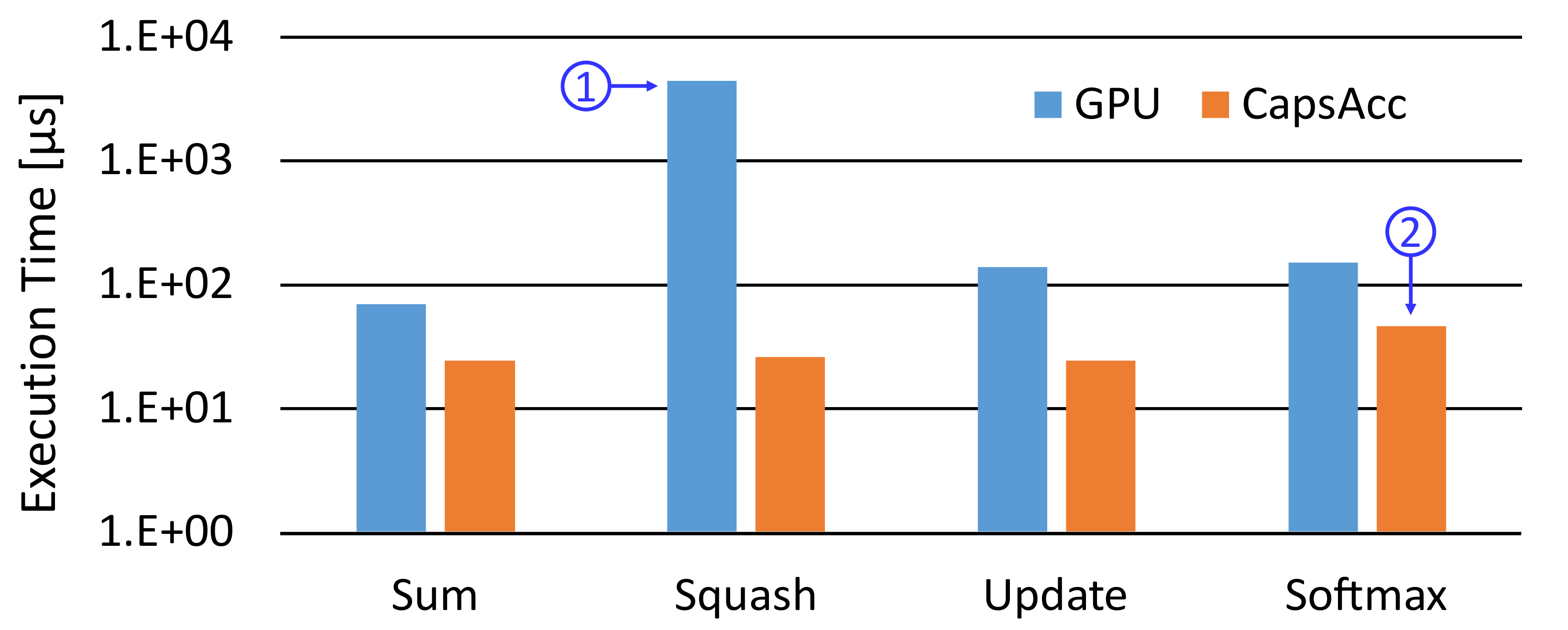}
    \vspace{-5pt}
    \caption{Execution time breakdown for the Dynamic Routing operations of the ShallowCaps~\cite{shallowcaps} on the Nvidia GeForce RTX 2080 Ti GPU and the CapsAcc hardware accelerator~\cite{Marchisio2019CapsAcc}.}
    \label{fig:motivation_figure}
    \vspace{-5pt}
\end{figure}

While for generic matrix multiplications (that are used on convolution operations) a common approach is to use approximate adders and multipliers~\cite{Marchisio2020ReD-CaNe}\cite{Mrazek2019ALWANN}, there are other complex operations (i.e., squash and softmax) that need more specialized designs to be computed in approximate form. Indeed, as shown in Fig.~\ref{fig:motivation_figure}, the squash and softmax operations are the most compute-intensive operations of the CapsNets. More precisely, \rpoint{1} the squash constitutes the performance bottleneck of their execution on GPUs, and \rpoint{2} the softmax has a high execution time on a Capsule Network Hardware Accelerator (CapsAcc~\cite{Marchisio2019CapsAcc}). Hence, these results motivate our research to focus on the design of approximate squash and softmax units. The research challenges tackled in this work are to find tradeoffs for area, power, and delay of the approximate softmax and squash units, without reducing the complete CapsNets' inference accuracy much.

\subsection{Novel Contributions}

Our novel contributions are:

\begin{itemize}[leftmargin=*]
    \item We analyze the state-of-the-art CapsNets models and the most advanced designs of approximate squash and approximate softmax (\textbf{Section~\ref{sec:background}}).
    \item We design specialized approximate softmax units using domain transformations (\textbf{Section~\ref{sec:Softmax_design}}).
    \item We design approximate squash units with piecewise approximations (\textbf{Section~\ref{sec:Squash_design}}).
    \item We implement the approximate softmax and squash architectures in VHDL, synthesize them in a 45nm technology node with the ASIC design flow, and perform gate-level simulations to evaluate the area, power consumption, and critical path delay.
    \item We also integrate the functional approximations into the open-source Q-CapsNets framework to evaluate the inference accuracy of state-of-the-art CapsNet models using the proposed approximate units (\textbf{Section~\ref{sec:results}}).
    \item Our proposed approximate \textit{softmax-b2} design outperforms the related works, having $-11\%$ area, $-8\%$ power, and $-19\%$ critical path delay, and comparable accuracy results.
    \item Our proposed approximate \textit{squash-exp} and \textit{squash-pow2} have up to $-6\%$ power consumption and up to $-36\%$ critical path delay compared to the state-of-the-art, while showing similar accuracy as having the exact squash function.
\end{itemize}

\section{Background and Related Works} 
\label{sec:background}

\subsection{Capsule Networks}
Capsule Networks (CapsNets) are introduced to improve the generalization ability of Convolutional Neural Networks (CNNs) in image classification tasks. 
CapsNets include multi-dimensional \textit{capsules}, i.e., groups of neurons that encode the existence probability and spatial properties of a specific feature, and exploit the dynamic \mbox{\textit{routing-by-agreement}} algorithm to detect entities that are consistent with lower-level features.

As a limitation, CapsNets show a higher computational complexity than traditional CNNs, as indicated by the MACs per memory ratio~\cite{qcaps}, due to the vectors of neurons and routing algorithm involving the iterative computation of complex operations, i.e., \textit{softmax} and \textit{squash}. 
The softmax function is a nonlinear function used to compute the routing coefficients connecting a lower-level capsule to the higher-level capsules. It normalizes its input values into a probability distribution. 
The squash function is the nonlinear activation function applied to produce the activity vector of the capsules. It ensures that the vector norm is below $1$ to represent the existence probability of the associated entity and preserves the vector orientation in agreement with the lower-level capsules predictions. 

The ShallowCaps model proposed by Hinton \textit{et al.}~\cite{shallowcaps} is designed for image classification on the \textit{MNIST} dataset~\cite{mnist} with greyscale images of handwritten digits. The architecture consists of three layers for the inference pass, which are a convolutional layer, a convolutional capsule layer, and a fully-connected capsule layer. The first layer has $256$, $9\times 9\times 1$ kernels with ReLU activation. The second layer applies $256$, $9\times 9\times 256$ kernels with a stride of $2$ and ReLU activation, and the output feature maps are reshaped in $32$ channels of $8$-dimensional capsules with squash activation. The final layer that performs the dynamic routing algorithm consists of $10$ $16$-dimensional capsules, one for each dataset class.

The \textit{DeepCaps} model~\cite{deepcaps} is introduced to improve the classification accuracy on complex image datasets like \mbox{\textit{CIFAR-10}}~\cite{cifar} with color images of animals and vehicles. The architecture consists of a convolutional layer, four middle stages (\textit{CapsCells}) including convolutional capsule layers (\textit{ConvCaps}) and a final fully-connected capsule layer. The two main architectural improvements w.r.t. the ShallowCaps are the skip connections that enable an efficient gradient flow during training and the routing-by-agreement algorithm based on 3D convolution to avoid the computational bottleneck that would occur by stacking multiple fully-connected capsule layers.

\subsection{Approximate Computing for DNNs Nonlinear Operations}
  
Approximate computing is an effective design methodology that aims to achieve low power consumption, high performance, and reduced circuit area by relaxing the accuracy requirement in error-tolerant applications~\cite{appcomp}. 
Extensive research efforts have been dedicated to the optimization of matrix multiplications in DNNs by proposing approximate designs for adders~\cite{appadd} and multipliers~\cite{appmpy}. 
However, a key factor for achieving high computing efficiency in DNNs and CapsNets is represented by the implementation of nonlinear functions, including nonlinearities such as sigmoid, hyperbolic tangent, softmax, and squash.

Various techniques have been proposed to compute nonlinear functions in an approximate form and enable an efficient hardware implementation with limited accuracy loss. 
The work in~\cite{appsigm} proposed a piecewise linear approximation of the sigmoid function by storing the curve breakpoints in a look-up table and applying linear interpolation.

As for the softmax function, the work in~\cite{taylor} proposed an approximate softmax design where the exponential function is evaluated by using Taylor series expansion and a look-up table method. The work in~\cite{lnu} presented a hardware architecture that exploits a mathematical transformation into the logarithmic domain to simplify the division operation and approximates the exponential and logarithmic functions by using linear fitting within a specific range. 

As regards the squash function, the work in~\cite{celebi} described a set of approximations of the Euclidean norm that avoid the computation of square and square-root operations. 
The work in~\cite{squashheal} introduced an approximate square-accumulate architecture with a self-healing mechanism that is suitable for computing the sum of squared components in the Euclidean norm. \textit{However, the previous works did not consider advanced methods, like piecewise approximations and domain transformations that are possible due to the error tolerance of these functions inserted in the CapsNets computations, that we indeed exploit in this work}.

\section{Approximate Softmax Design} 
\label{sec:Softmax_design}

\begin{figure*}[ht]
\vspace{0pt}
\centering
\subfloat[][\label{fig:1a}]{\includegraphics[scale=0.4]{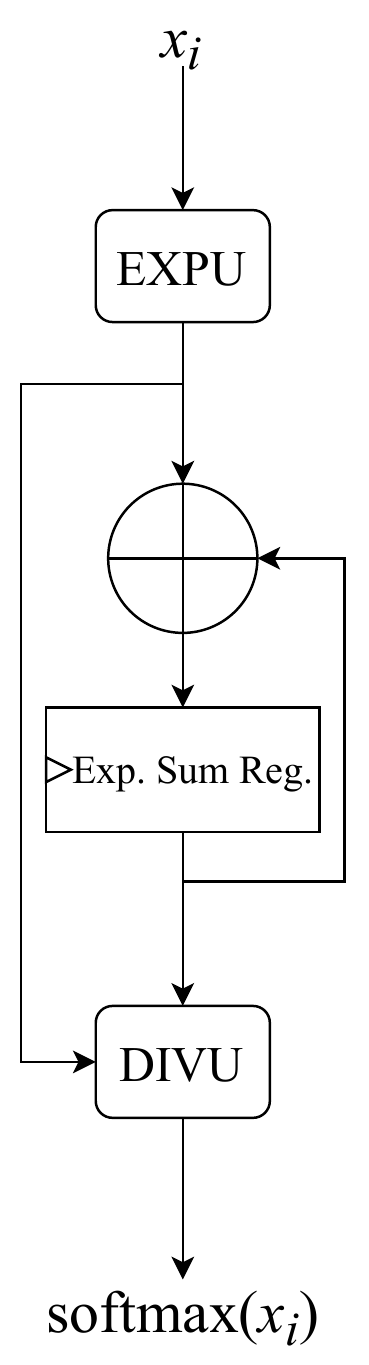}} \quad
\subfloat[][\label{fig:1b}]{\includegraphics[scale=0.4]{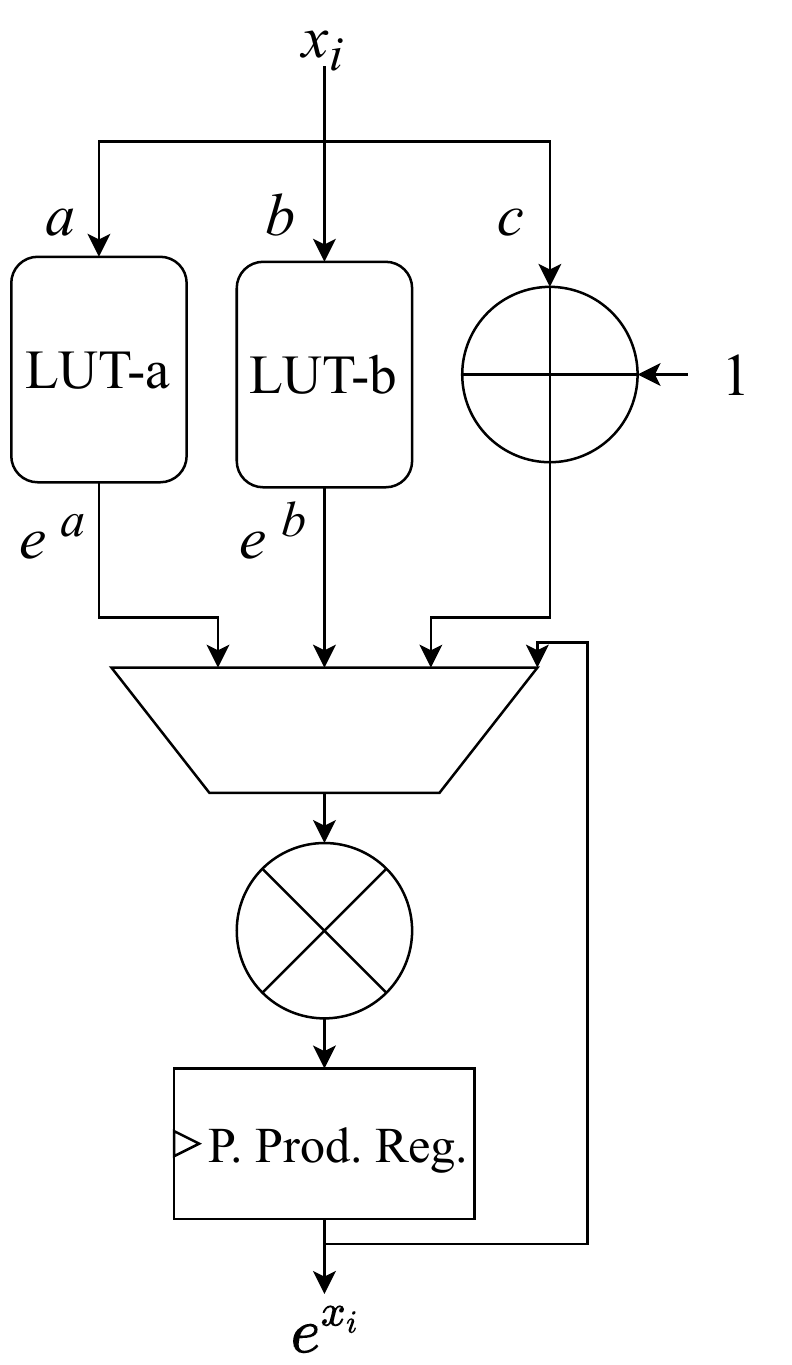}} \quad
\subfloat[][\label{fig:1c}]{\includegraphics[scale=0.4]{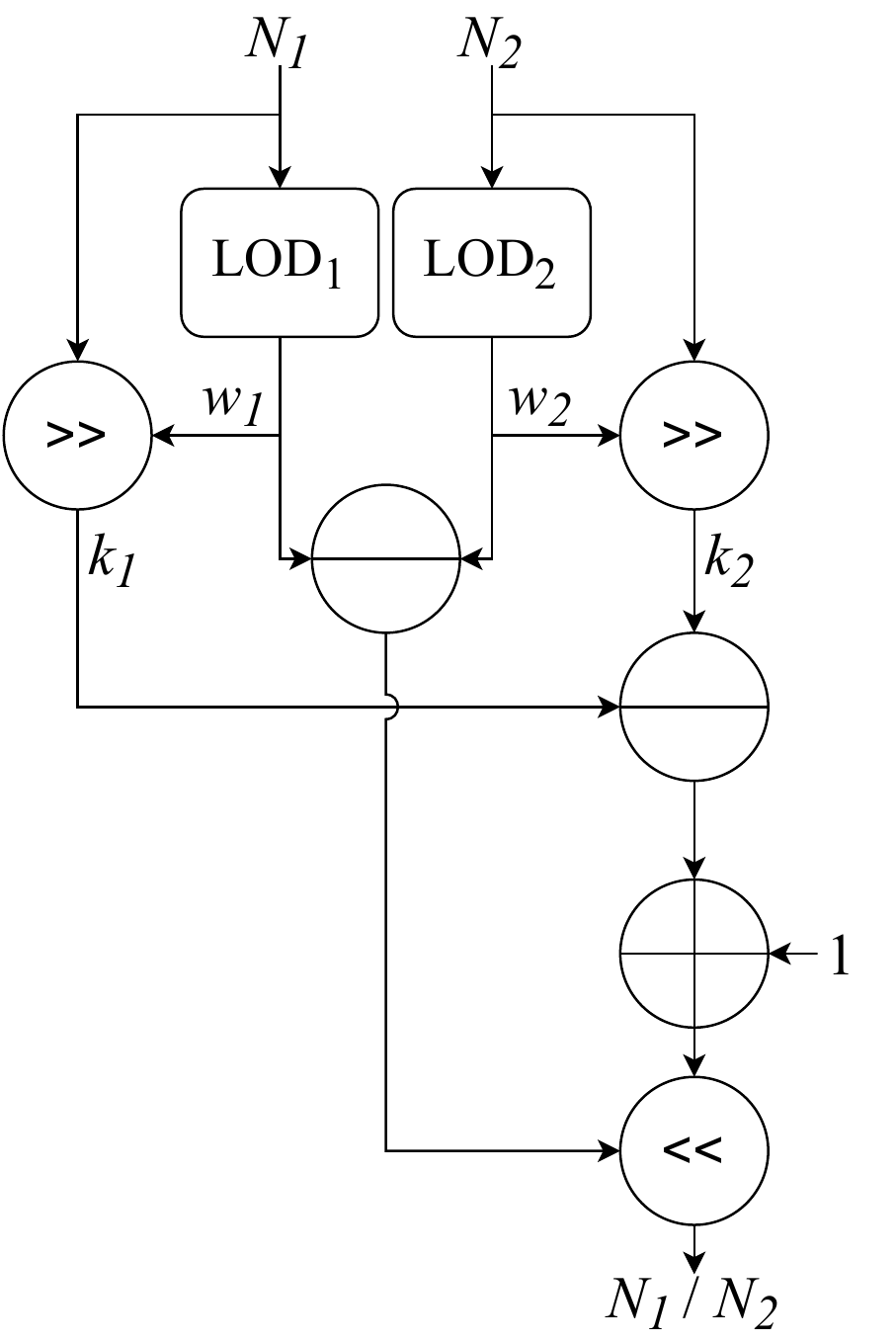}} \quad
\subfloat[][\label{fig:1d}]{\includegraphics[scale=0.4]{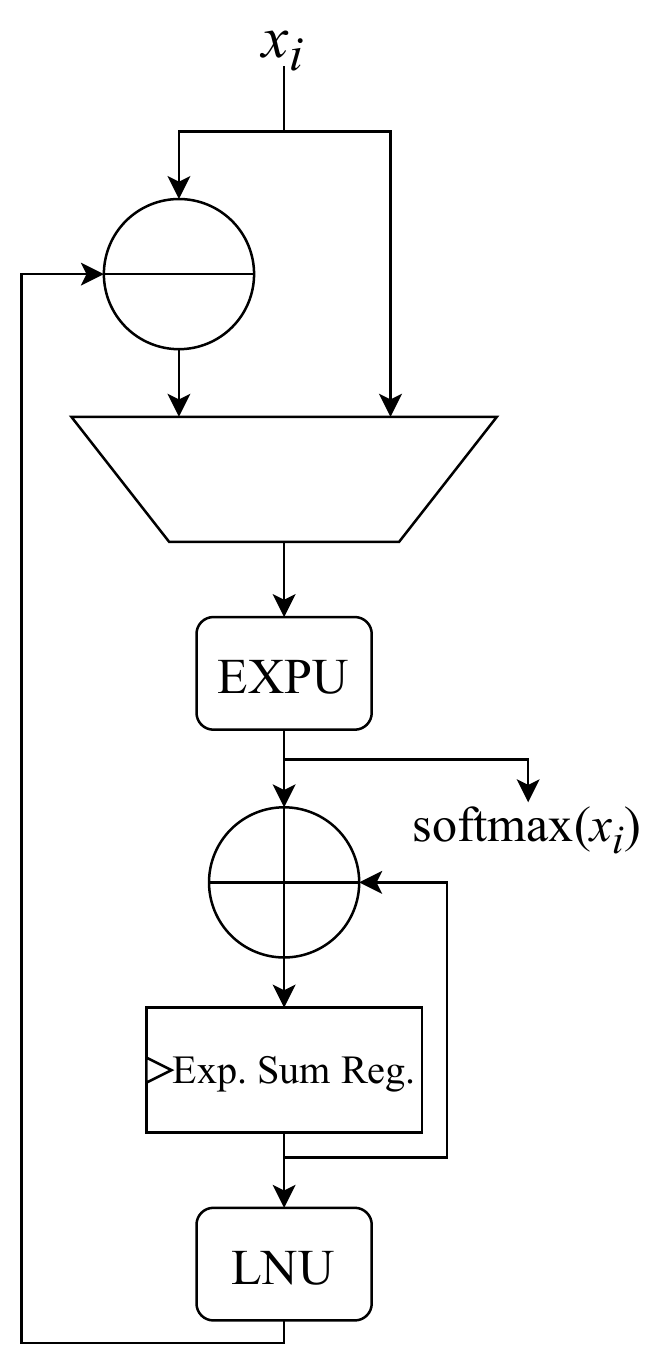}} \quad
\subfloat[][\label{fig:1e}]{\includegraphics[scale=0.4]{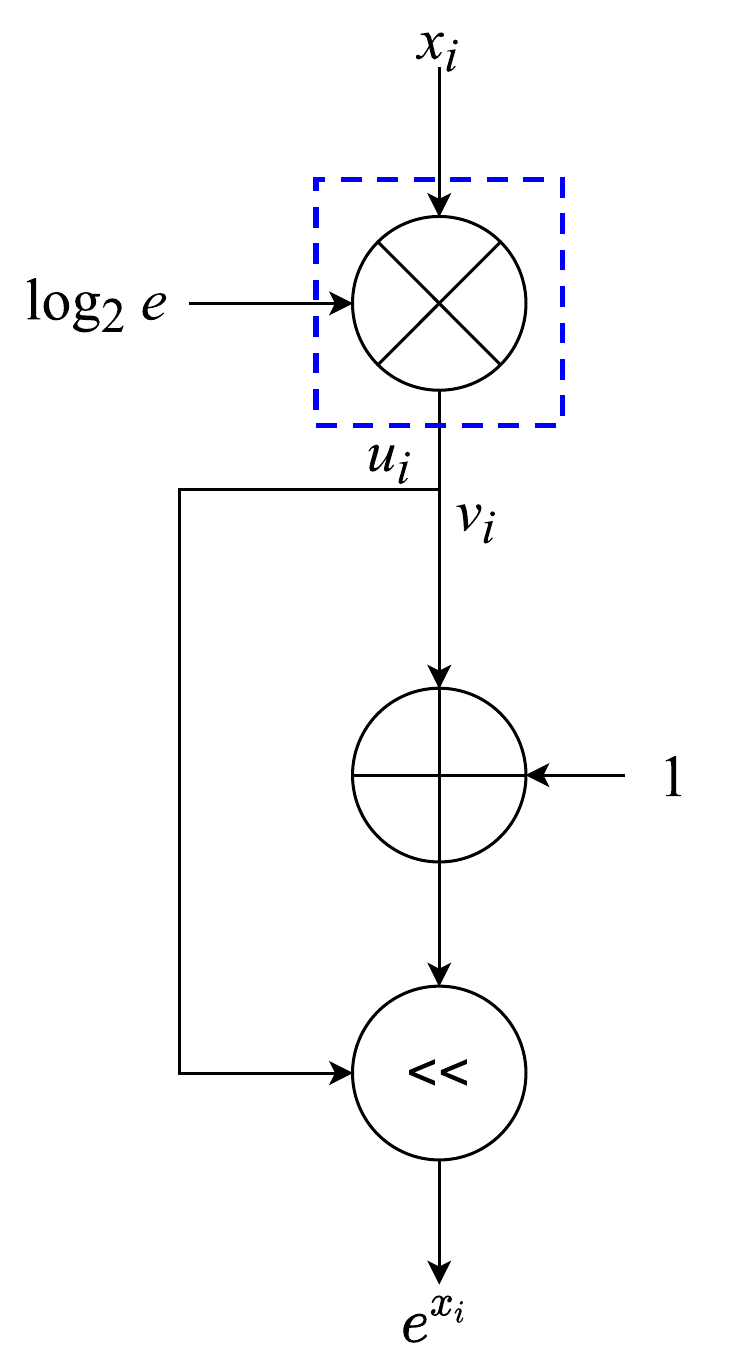}} \quad
\subfloat[][\label{fig:1f}]{\includegraphics[scale=0.4]{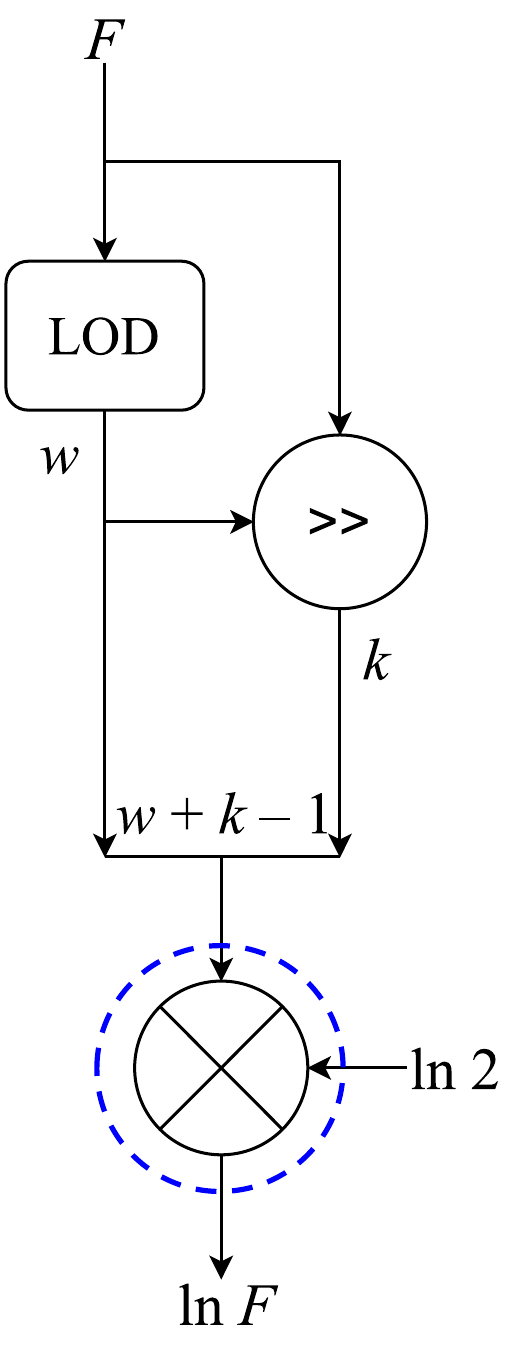}}
\vspace{-5pt}
\caption{Architectures of the approximate softmax designs: (a) Softmax-taylor unit. (b) Softmax-taylor exponent unit. (c) Softmax-taylor division unit. (d) Softmax-lnu unit. (e) Softmax-lnu exp unit. (f) Softmax-lnu natural log unit. Softmax-b2 unit replaces expu and lnu in Softmax-lnu with pow2u and log2u by removing the multipliers $\log_2 e$ and $\ln 2$.}
\label{fig:1}
\vspace{-5pt}
\end{figure*}

In the following, we present three approximate softmax designs describing the algorithmic approximations and the RTL architectures. The proposed softmax approximations are referred to as \textit{softmax-taylor}, \textit{softmax-lnu} and \textit{softmax-b2}, with names enclosing their key features. 

The softmax function shown in Eq.~\ref{eq:softmax} is a probabilistic version of the argmax function, which returns $1$ for the highest input value and $0$ for all the other values.
\begin{equation}
\label{eq:softmax}
y_i=\frac{e^{x_i}}{\sum_{j=1}^n e^{x_j}} \quad (i=1,...,n)
\end{equation}

The softmax computation involves three fundamental operations: natural exponential, sum, and division. In the following approximate softmax designs, we mainly focus on the approximate computation of the exponentiation and division, which are the most complex operations of the softmax function.

The \textbf{softmax-taylor} design is based on a specific softmax approximation~\cite{taylor} which exploits the Taylor series expansion method for exponential computation and performs divisions in the logarithmic domain. 

The natural exponent operation is simplified as in Eq.~\ref{eq:exp_taylor} using the first-order Taylor polynomial approximation. At the architecture level, the exponent unit consists of $2$ look-up tables to implement the first two exponent contributions, a specific bus arrangement to get $1+c$ and a multiplier to compute the final product iteratively (see~\figurename\,\ref{fig:1a} and \figurename\,\ref{fig:1b}).

\begin{equation}
\label{eq:exp_taylor}
e^{x_i}=e^{a+b+c}\approx e^{a}\cdot e^{b}\cdot (1+c)
\end{equation}

The division operation is performed in the logarithmic domain by exploiting the mathematical transformation in Eq.~\ref{eq:pow2} 

\begin{align}
\label{eq:pow2}
\begin{split}
&\resizebox{\linewidth}{!}{\ensuremath{\text{pow2}\, (\log_2\, (e^{x_i}/ \sum_{j=1}^n e^{x_j}))=\text{pow2}\, (w_1 + \log_2 k_1 - (w_2 + \log_2 k_2))}}\\
&\resizebox{.8\linewidth}{!}{\ensuremath{\approx \text{pow2}\, (w_1 - w_2 + k_1 - k_2))=2^{u_i+v_i}\approx 2^{u_i}\cdot (1+v_i)}}
\end{split}
\end{align}

First of all, $N_1=e^{x_i}$ and $N_2=\sum_{j=1}^n e^{x_j}$ are expressed as $2^{w_l}\cdot k_l$, with $w_l \in \mathbb{Z}$ and $k_l \in [1, 2)$ for $l=1,2$ and the base-2 logarithm of $k_l$ is approximated by the linear fitting function $k_l-1$. Secondly, the argument of the power-2 operation is split into its integer and fractional parts, $u_i$ and $v_i$, with $u_i \in \mathbb{Z}$ and $v_i \in [0, 1)$ and $2^{v_i}$ is estimated as $(1+v_i)$.

The division unit is composed of $2$ base-2 logarithm units, a leading one detector (LOD) and shift unit that compute the logarithm of the dividend and divisor, a subtraction unit that performs the division in the log domain, and a power-2 unit (bus arrangement and shift unit) that computes the softmax output value (see~\figurename\,\ref{fig:1c}).

To be compliant with the capsule network models~\cite{shallowcaps}~\cite{deepcaps} that we use in our experiments, the softmax architecture is able to process $10$, $32$ or $128$ inputs.

The \textbf{softmax-lnu} design builds on a peculiar softmax approximation~\cite{lnu} which adopts a mathematical domain transformation involving natural logarithm and natural exponential operations (see Eq.~\ref{eq:exp_lnu}).
\begin{equation}
\label{eq:exp_lnu}
\text{exp}\, (\ln\, (e^{x_i}/ \sum_{j=1}^n e^{x_j}))=\text{exp}\, (x_i - \ln\, (\sum_{j=1}^n e^{x_j}))
\end{equation}

The transformation into the logarithm domain allows to perform the division by using a more straightforward subtraction. On the other hand, the exponentiation is required to convert the softmax output values from the logarithmic domain to the linear one. The design architecture mainly consists of three computational units to compute the natural exponential of the softmax inputs (EXPU), sum up the exponentials, and evaluate the natural logarithm of the sum (LNU) required for the division (see~\figurename\,\ref{fig:1d}).

The natural exponential operation is performed by using the mathematical transformation in Eq.~\ref{eq:exp_lnu2}, with $u_i \in \mathbb{Z}$ and $v_i \in [0, 1)$. At the architecture level, the natural exponential unit is composed of a constant multiplier by $\log_2 e$, a specific bus arrangement to implement $1+v_i$ and a shift unit to compute the result (see~\figurename\,\ref{fig:1e}).
\begin{equation}
\label{eq:exp_lnu2}
e^{x_i}=2^{x_i\cdot \log_2 e}=2^{u_i+v_i}=2^{u_i}\cdot 2^{v_i}\approx 2^{u_i}\cdot (1+v_i)
\end{equation}

The natural logarithm is computed as in Eq.~\ref{eq:ln_lnu}, where $F=\sum_{j=1}^n e^{x_j}$ is expressed as $2^w\cdot k$, with $w \in \mathbb{Z}$ and $k \in [1, 2)$ and the base-2 logarithm of $k$ is approximated by the linear fitting function $k-1$.  The natural logarithm unit consists of $4$ main subunits: a leading one detector to determine $w$, a shift unit to compute $k$, a specific bus arrangement to get the base-2 logarithm of $F$ and a constant multiplier by $\ln 2$ (see~\figurename\,\ref{fig:1f}).
\begin{equation}
\label{eq:ln_lnu}
\ln F = \ln 2\cdot \log_2 F = \ln 2\cdot (w + \log_2 k) \approx \ln 2\cdot (w + k - 1)
\end{equation}

The architecture includes other hardware units to compute the maximum input value, scale the inputs, execute the division in the log domain, and allow for the processing of a variable number of softmax inputs. 

The \textbf{softmax-b2} design implements the idea of computing a softmax-like function with powers of 2 in place of natural exponentials, and it exploits a domain transformation with base-2 logarithm and power-2 operations (see Eq.~\ref{eq:b2}).
\begin{equation}
\label{eq:b2}
\text{pow2}\, (\log_2\, (2^{x_i}/ \sum_{j=1}^n 2^{x_j}))=\text{pow2}\, (x_i - \log_2\, (\sum_{j=1}^n 2^{x_j}))
\end{equation}

The proposed approximation allows for a complexity reduction of the hardware implementation of the \textit{softmax-lnu} design, thanks to the removal of two constant multipliers. 

\begin{figure*}[ht]
\vspace{0pt}
\centering
\subfloat[][\label{fig:2a}]{\includegraphics[scale=0.4]{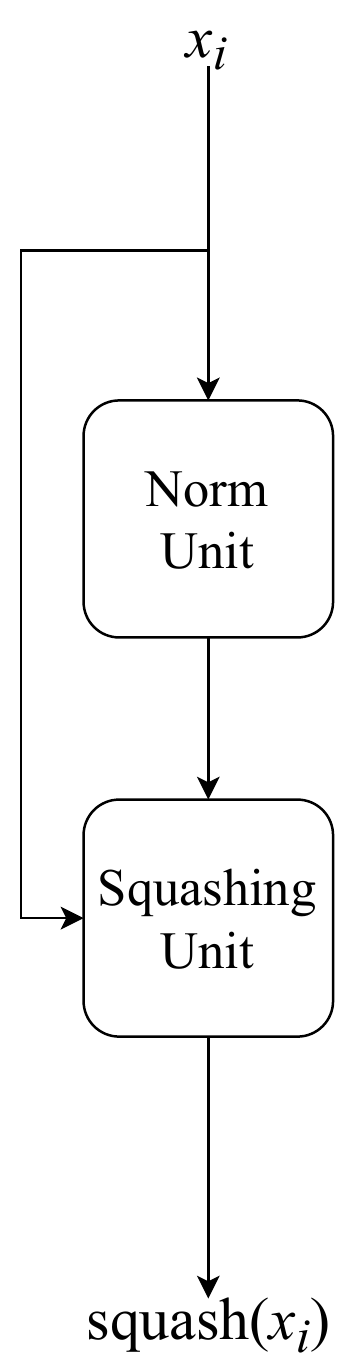}} \qquad
\subfloat[][\label{fig:2b}]{\includegraphics[scale=0.4]{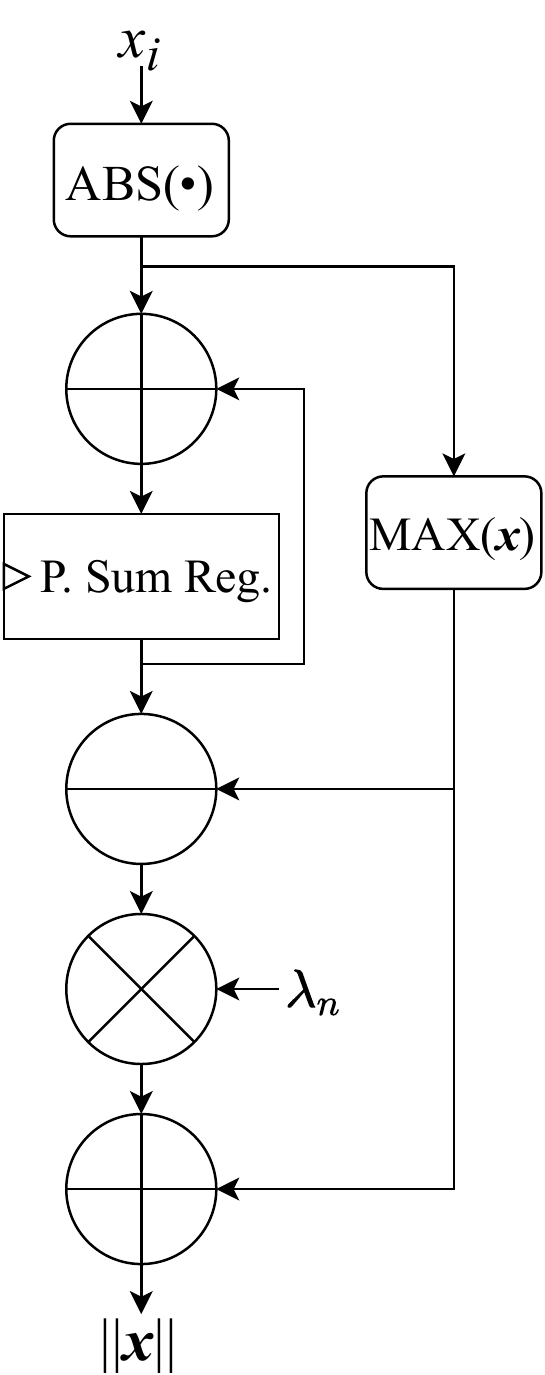}} \qquad
\subfloat[][\label{fig:2c}]{\includegraphics[scale=0.4]{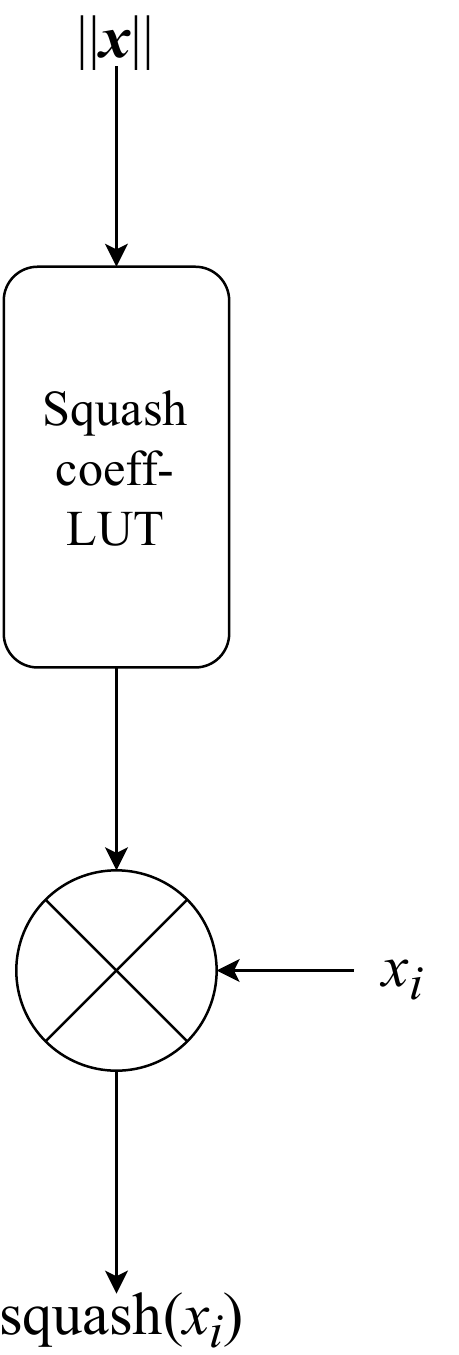}} \qquad
\subfloat[][\label{fig:2d}]{\includegraphics[scale=0.4]{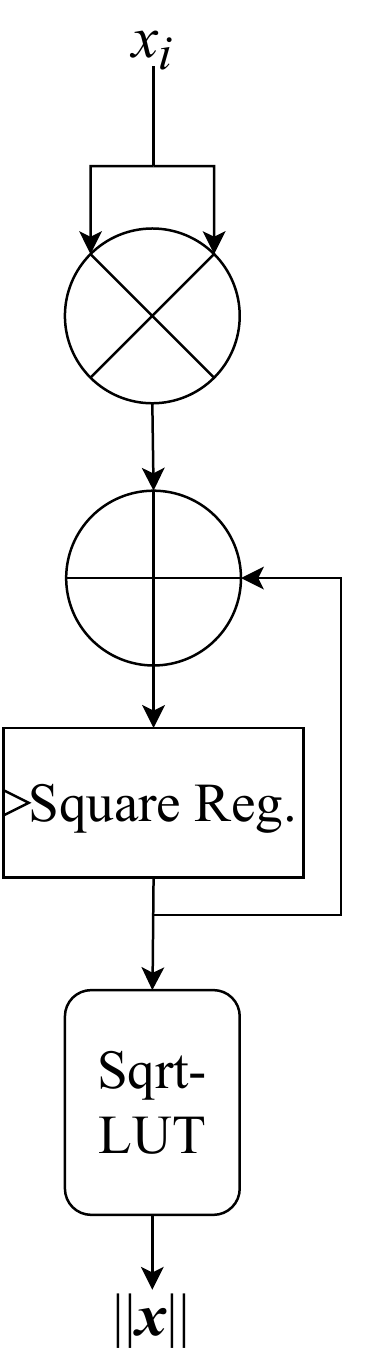}} \qquad
\subfloat[][\label{fig:2e}]{\includegraphics[scale=0.4]{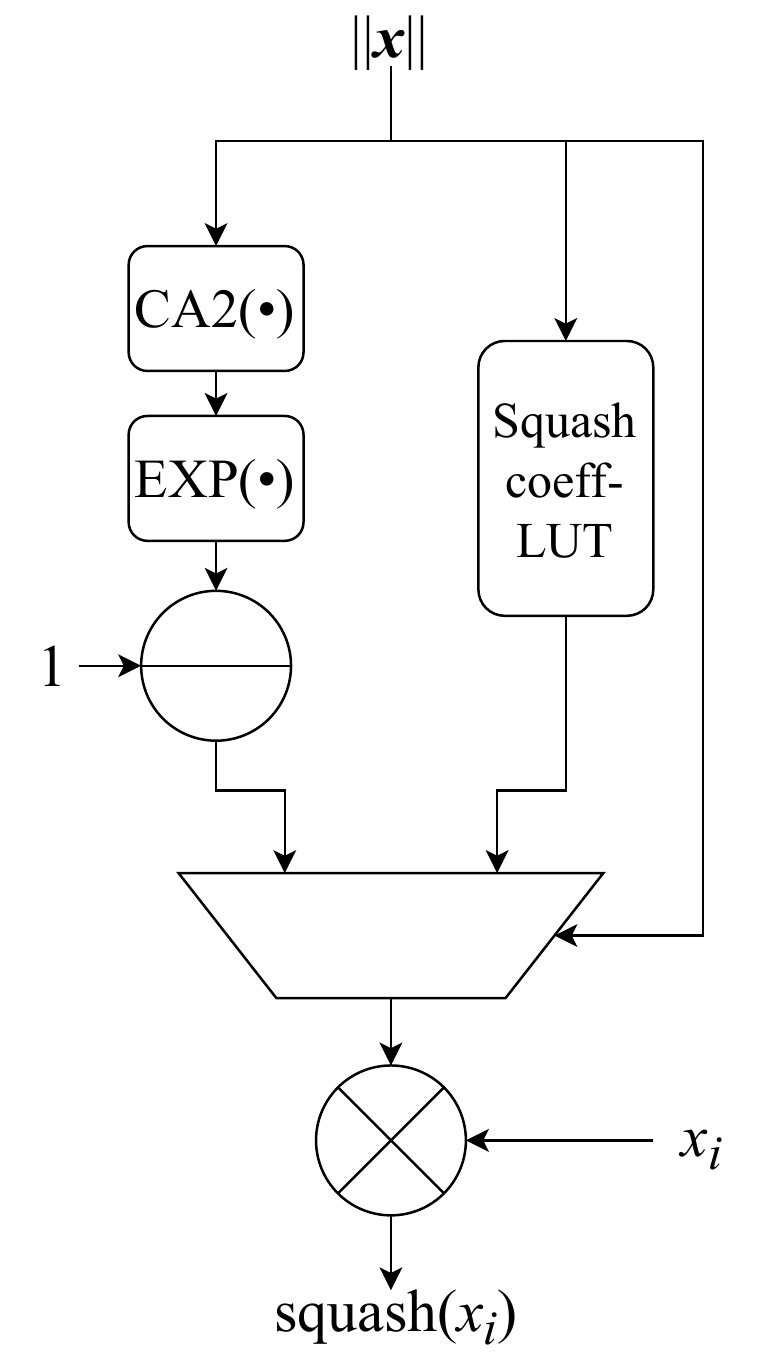}} \qquad
\subfloat[][\label{fig:2f}]{\includegraphics[scale=0.4]{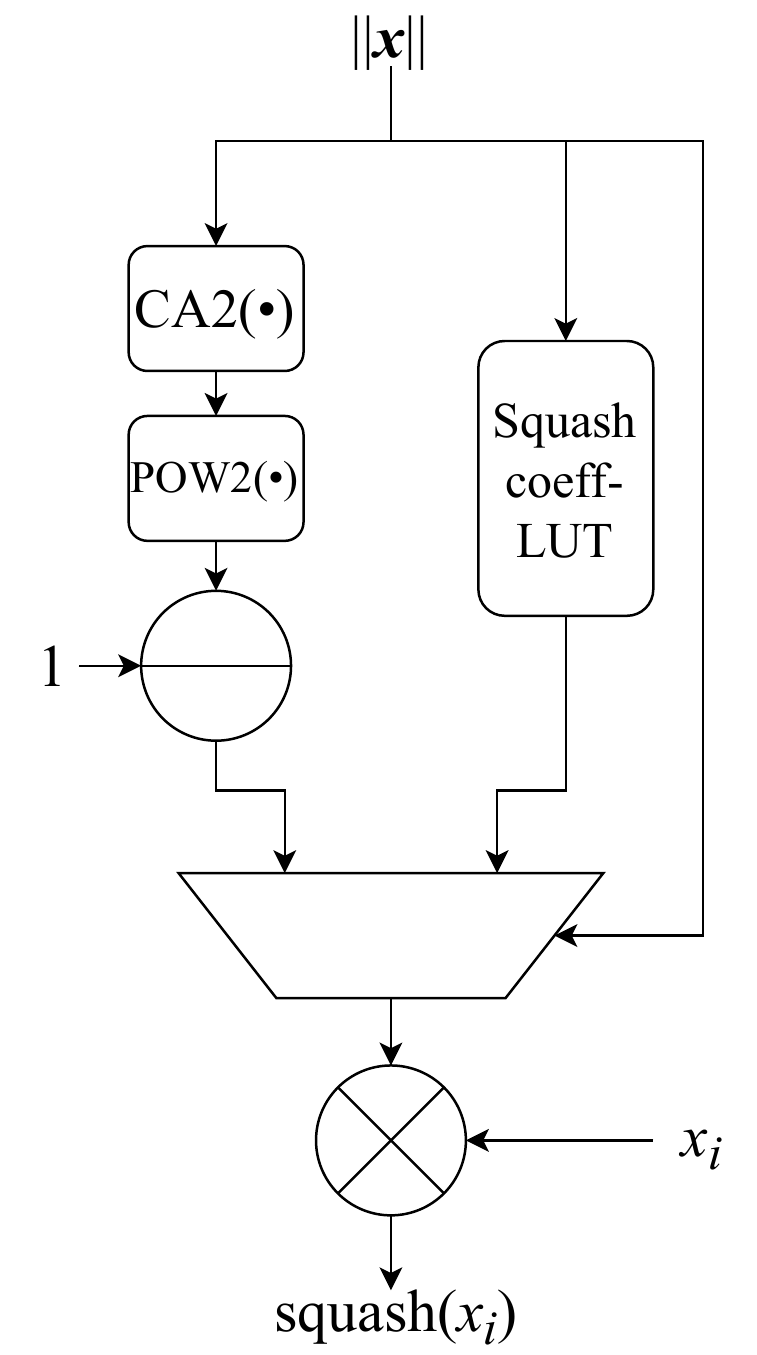}}
\vspace{-5pt}
\caption{Architectures of the approximate squash designs: (a) Squash function unit. (b) Squash-norm norm unit. (c) Squash-norm squashing unit. (d) Squash-exp and -pow2 norm unit. (e) Squash-exp squashing unit. (f) Squash-pow2 squashing unit.}
\label{fig:2}
\vspace{-5pt}
\end{figure*}

Compared to the \textit{softmax-lnu} design, the \textit{softmax-b2} architecture avoids the preliminary multiplication by $\log_2 e$ in the exponential unit (see dashed square in~\figurename\,\ref{fig:1e}) and the final multiplication by $\ln 2$ in the logarithmic unit (see dashed circle in~\figurename\,\ref{fig:1f}), by implementing the power-2 and base-2 logarithm unit, respectively.

\section{Approximate Squash Design} 
\label{sec:Squash_design}

The proposed approximate squash designs are called \textit{squash-norm}, \textit{squash-exp} and \textit{squash-pow2}.

The squash function requires to compute the norm of the input vector and the squashing coefficient that multiplies the input vector to produce the output vector, as shown in Eq.~\ref{eq:squash}. 
\begin{equation}
\label{eq:squash}
\mathbf{y}=\frac{\norma{\mathbf x}^2}{1+\norma{\mathbf x}^2}\, \frac{\mathbf x}{\norma{\mathbf x}}
\end{equation}

The first design exploits a specific norm approximation~\cite{celebi}, while the remaining two techniques introduce novel solutions to approximate the squashing coefficient.

The \textbf{squash-norm} design is inspired by the specific Euclidean norm approximation proposed by Chaudhuri \mbox{\textit{et al.}}~\cite{celebi}, which is shown in Eq.~\ref{eq:appnor}. 
\begin{equation}
\label{eq:appnor}
\norma{\mathbf{x}} \approx D_{\lambda}(\mathbf{x}) = \abs {x_{i_{max}}} + \lambda\! \sum_{\substack{i=1\\ i\ne i_{max}}}^n \abs {x_i}
\end{equation}

This architecture does not require the square root operator and the multiplications needed to square the vector components, but it involves the computation of the absolute values and the maximum absolute value components. The parameter $\lambda$ depends on the number of vector components and it is selected accordingly~\cite{rhodes}.

The designed architecture is composed of two main units. The norm unit computes the approximate vector norm, and the squashing unit produces the squash outputs (see~\figurename\,\ref{fig:2a}).

The norm unit implements the Chaudhuri approximation~\cite{celebi} in Eq.~\ref{eq:appnor}. It consists of multiple arithmetic modules. A dedicated component computes the absolute value of the inputs, an accumulator sums up the absolute values, a unit determines the maximum absolute value, a subtractor gets the second term of the formula, a multiplier scales the sum by $\lambda$ and an adder adds the maximum value to the sum (see~\figurename\,\ref{fig:2b}).

The squashing unit consists of two look-up tables to implement the squashing coefficient and a multiplier to compute the squash outputs as the product between the inputs and the squashing coefficient (see~\figurename\,\ref{fig:2c}).

To be compliant with the two capsule network models employed in our experiments, the squash architecture is able to process $4$, $8$, $16$, or $32$ inputs.

The \textbf{squash-exp} design exploits a piecewise approximation of the squashing coefficient $\norma{\mathbf x}/(1+\norma{\mathbf x}^2)$ in two ranges of norm values. The coefficient is approximated by the nonlinear function $1-e^{-\norma{\mathbf x}}$ in the first range and by a direct mapping method in the second range. The range of norm values is derived experimentally by executing inference steps with two capsule network models on two image datasets.

At the architecture level, the design mainly consists of two computational units: the norm unit and the squashing unit. 

The norm unit computes the Euclidean norm of the input vector. It is composed of a multiplier to square the input components, an accumulator to sum up the squared inputs, and two look-up tables to implement the square root function over two specific ranges of squared norm values (see~\figurename\,\ref{fig:2d}).

The squashing unit implements the piecewise approximation of the squashing coefficient and computes the output values. The nonlinear function in the first range is implemented by a component composed of a 2's complement of the norm value, a natural exponential unit, and a subtractor. The second-range approximation is performed with a look-up table. The final multiplier is used to compute the squash outputs (see~\figurename\,\ref{fig:2e}).

The \textbf{squash-pow2} design builds on the piecewise approximation of the squashing coefficient used in the \textit{squash-exp} architecture, but the approximating nonlinear function used in the first range of norm values is $1-2^{-\norma{\mathbf x}}$.

\begin{figure}[ht]
\centering
\vspace{-6pt}
\subfloat[][]{\includegraphics[width=0.47\linewidth]{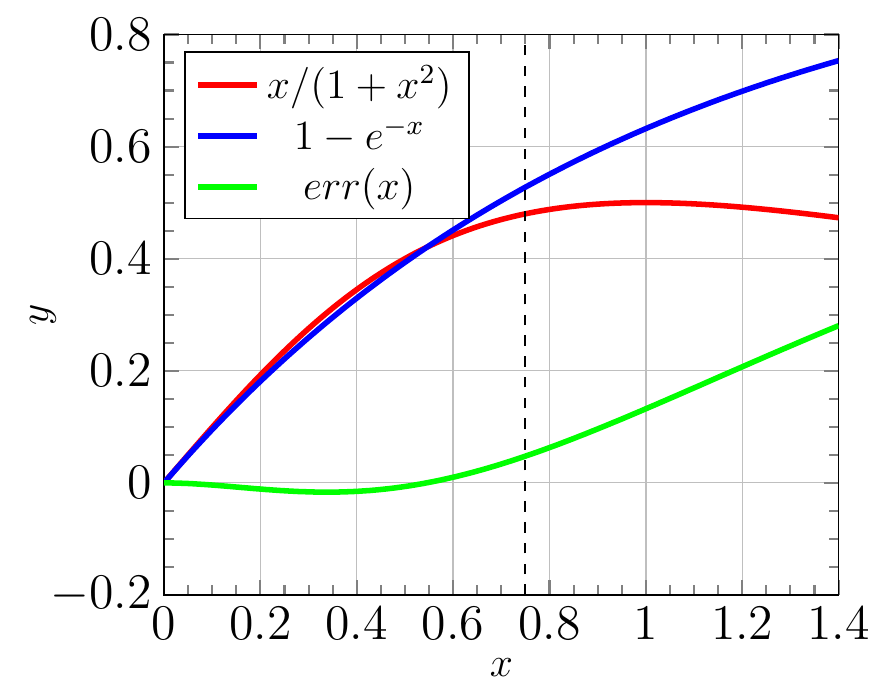}} \quad
\subfloat[][]{\includegraphics[width=0.47\linewidth]{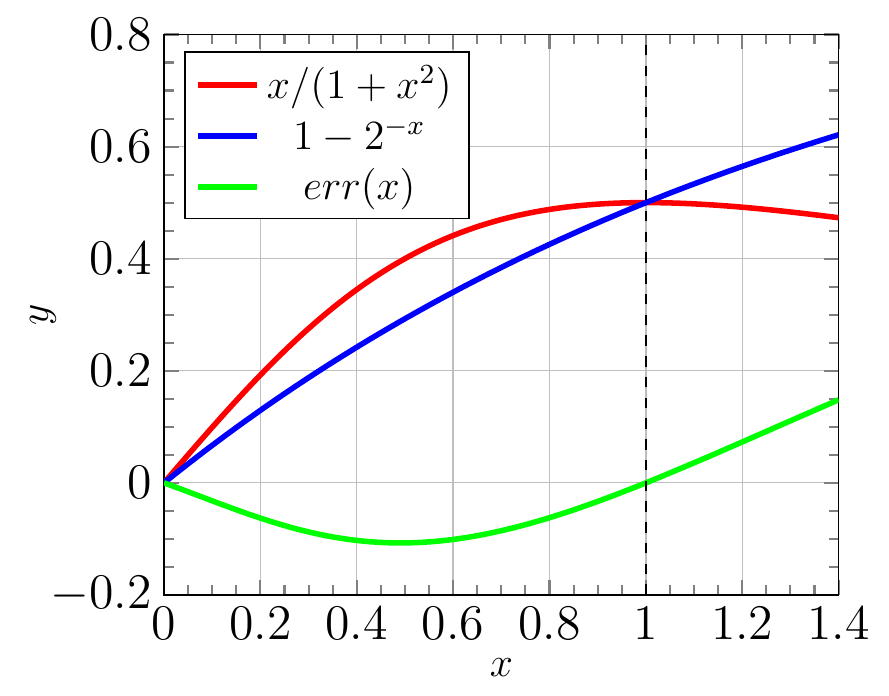}}
\vspace{-9pt}
\caption{Behavior of (a) squash-exp and (b) squash-pow2 approximation with $x\vcentcolon =\lVert \mathbf{x}\rVert$.}
\label{fig:3}
\vspace{-12pt}
\end{figure}

At the architecture level, in the exponential unit the constant multiplication by $\log_2 e$ is removed to implement the power-2 unit (see~\figurename\,\ref{fig:2f}). The hardware cost reduction is obtained at the expense of a higher worst-case approximation error of the squashing coefficient in the range of low norm values (see~\figurename\,\ref{fig:3}).

\section{Evaluation of our Designs} 
\label{sec:results}

In the following, the approximate softmax and squash designs are evaluated in terms of inference accuracy loss and hardware implementation metrics. 

First, we explore the inference accuracy degradation induced by the proposed softmax and squash approximations in $4$ case studies, with two capsule network models on two image classification datasets. 
Secondly, we synthesize the complete architectures and analyze our designs' area usage, power consumption, and timing performance. 

The objective of the evaluation is to explore possible \textit{trade-offs} between the classification accuracy loss of a CapsNet using the approximations and the hardware implementation cost of the approximate designs.

\subsection{Experimental Setup}
We implement the approximate softmax and squash algorithms in Python and perform extensive software simulations to evaluate the quality of each approximation w.r.t. the exact function. The experiments are conducted for over $1,\!000$ input vectors in a specific range. We analyze the \textit{Mean Error Distance} on the maximum and average component errors, in absolute and relative terms.

To assess how the softmax and squash approximations affect the inference accuracy of the complete capsule networks, we include the approximate functions in a Python-based CapsNet model provided by the open-source framework \textit{Q-CapsNets}~\cite{qcaps} and we perform an image classification task with two CapsNet models, ShallowCaps~\cite{shallowcaps} and DeepCaps~\cite{deepcaps}, on two image datasets, MNIST~\cite{mnist} and Fashion-MNIST~\cite{fashionmnist}. 

\begin{figure}[h]
    \centering
    \includegraphics[width=.99\linewidth]{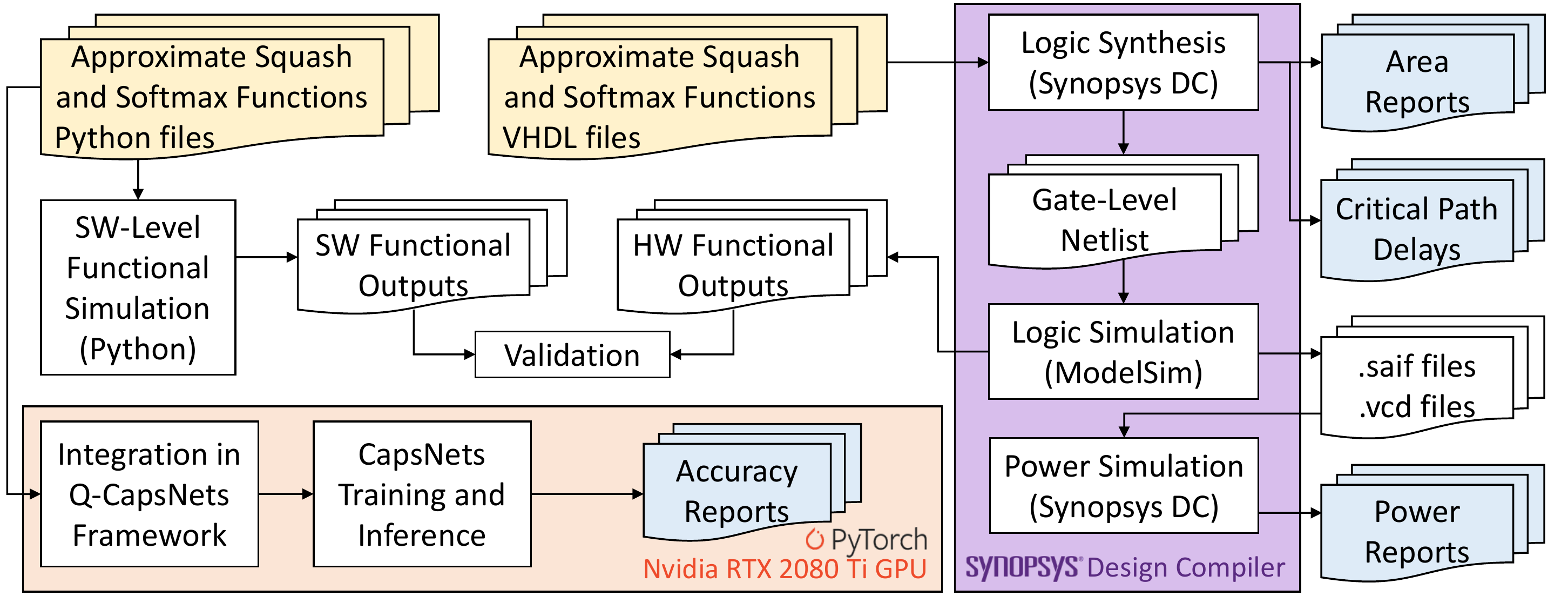}
    \vspace{-6pt}
    \caption{Setup and tool-flow for conducting our experiments.}
    \label{fig:exp_setup}
    \vspace{-12pt}
\end{figure}

As shown in~\figurename\,\ref{fig:exp_setup}, our experimental setup consists of both software and hardware components. We use a software environment with PyTorch library and Nvidia CUDA Toolkit and execute the inference passes on an Nvidia GeForce RTX 2080 Ti GPU. To comply with the hardware implementation, we perform the quantization of the approximate softmax and squash data, and we test the quantized approximate functions in quantized CapsNet models (see~\tablename\,\ref{tab:1}). Using the Q-CapsNets framework, we quantize weights and activations of the CapsNet models on the image datasets and input data of the softmax and squash functions.

\begin{table}[ht]
\caption{Percentage of \textit{quantized} inference accuracy.}
\vspace{-9pt}
\label{tab:1}
\centering
\resizebox{\columnwidth}{!}{
\begin{tabular}{@{}lcccc@{}}
\toprule
{} & \multicolumn{2}{c}{MNIST} & \multicolumn{2}{c}{Fashion-MNIST} \\
\cmidrule(lr){2-3}
\cmidrule(l){4-5}
{} & ShallowCaps & DeepCaps & ShallowCaps & DeepCaps \\
\midrule
\textit{exact functions}	& 99.44 & 99.35 & 92.42 & 94.69 \\
\midrule
\textbf{softmax-lnu}		& 99.46 & \textbf{99.42} & 92.37 & \textbf{94.71} \\
\textbf{softmax-b2 (ours)}		& \textbf{99.49} & 99.33 & 92.33 & 94.64 \\
\textbf{softmax-taylor}	& 99.42 & 99.41 & \textbf{92.47} & 94.69 \\
\midrule
\textbf{squash-exp (ours)}		& 99.18 & 98.79 & 91.32 & \textbf{94.76} \\
\textbf{squash-pow2 (ours)}		& 99.00 & 98.58 & 89.05 & 94.62 \\
\textbf{squash-norm}		& \textbf{99.26} & \textbf{99.23} & \textbf{92.51} & 94.70 \\
\bottomrule
\end{tabular}
}
\vspace{-2pt}
\end{table}

We implement the design architectures in VHDL and perform functional simulation using ModelSim to check the results against the Python model outputs. We synthesize the architectures in a 45nm academic technology library, \textit{Nangate OCL}, by using the ASIC design flow with Synopsys Design Compiler and obtain area usage, power consumption, and maximum path delay of each design (see~\tablename\,\ref{tab:2}). Finally, we conduct post-synthesis functional and timing validation of the gate-level netlist.

\begin{table}[ht]
\vspace{-6pt}
\caption{Hardware characteristics with clock frequency $100$\,MHz.}
\vspace{-9pt}
\label{tab:2}
\centering
\resizebox{\columnwidth}{!}{
\begin{tabular}{@{}lccc@{}}
\toprule
{} & Area usage & Power consumption & Critical path delay \\
{} & ($\mu m^2$) & ($\mu W$) & ($ns$) \\
\midrule
\textbf{softmax-lnu}		& 12,511 & 2,572 & 6.46 \\
\textbf{softmax-b2 (ours)}		& \textbf{11,169} & \textbf{2,244} & \textbf{4.22} \\
\textbf{softmax-taylor	}	& 14,944 & 2,430 & 5.24 \\
\midrule
\textbf{squash-exp (ours)}		& 7,937 & 1,414 & 5.64 \\
\textbf{squash-pow2 (ours)}		& 7,543 & \textbf{1,340} & \textbf{4.17} \\
\textbf{squash-norm}		& \textbf{6,806} & 1,431 & 6.53 \\
\bottomrule
\end{tabular}
}
\vspace{-7pt}
\end{table}

\subsection{Evaluating the Softmax}
\label{subsec:Softmax_eval}

From the experimental results, we derive the following key observations regarding the approximate softmax designs.

The \textbf{softmax-b2} design is the best solution in terms of hardware metrics but it implies the highest CapsNet accuracy loss in all the case studies except for the ShallowCaps on MNIST. Actually, the \textit{b2} design consumes less area ($-11\%$ and $-25\%$) and power ($-13\%$ and $-8\%$) than the \textit{lnu} and \textit{taylor} designs. Moreover, it has the lowest critical path delay ($-35\%$ and $-19\%$ w.r.t. \textit{lnu} and \textit{taylor}). 

The \textbf{softmax-taylor} design is the best choice in terms of inference accuracy loss since it outperforms the other designs in the ShallowCaps for Fashion-MNIST. However, it is characterised by the worst area usage ($+20\%$ and $+35\%$ w.r.t. \textit{lnu} and \textit{b2}) and intermediate power consumption and critical path delay.

The \textbf{softmax-lnu} design shows the highest power consumption ($+15\%$ and $+5\%$ w.r.t. \textit{b2} and \textit{taylor}) and maximum path delay ($+53\%$ and $+23\%$) but intermediate area usage. Its performance in inference accuracy loss is similar to the \textit{taylor} design in all the case studies, except for the ShallowCaps for Fashion-MNIST, where the \textit{lnu} performs worse ($+0.1\%$ loss).

\subsection{Evaluating the Squash}
\label{sec:Squash_eval}

The \textbf{squash-norm} design is the best approximate squash solution in terms of CapsNet accuracy loss. It also has the benefit of having the best area usage ($-13\%$ and $-8\%$ w.r.t. \textit{exp} and \textit{pow2}), but as a drawback, it shows the worst power ($+1\%$ and $+7\%$) and delay metrics ($+15\%$ and $+56\%$).

The \textbf{squash-pow2} design is the best option in terms of power consumption ($-5\%$ and $-6\%$ w.r.t. \textit{exp} and \textit{norm}) and critical path delay ($-25\%$ and $-36\%$), and it has intermediate area usage. However, it implies the highest CapsNet accuracy loss among all the case studies.

The \textbf{squash-exp} design is characterized by an accuracy loss similar to the \textit{norm} design in two case studies and significantly worse accuracy in the other two cases. In exchange for the reduced accuracy, it has intermediate power and delay metrics, but as a downside, it shows the worst area usage ($+5\%$ and $+17\%$ w.r.t. \textit{pow2} and \textit{norm}).

\section{Conclusion} 
To enable efficient CapsNets inference on edge devices, we propose approximate designs for the most compute-intensive CapsNets operations, which are the softmax and squash. Our \textit{softmax-b2} design based on approximating the natural exponential with powers of 2 significantly reduces the hardware complexity, with limited accuracy drop. Our squash designs based on piecewise approximations show interesting tradeoffs between accuracy, area, power consumption, and critical path delay. We believe that our findings will contribute to the deployment of CapsNets and other complex DNN models on resource-constrained devices.


\begin{acks}
This work has been supported in part by the Doctoral College Resilient Embedded Systems, which is run jointly by the TU Wien's Faculty of Informatics and the UAS Technikum Wien. 
\end{acks}

\bibliographystyle{ACM-Reference-Format}
\bibliography{main}


\end{document}